\documentstyle[12pt]{article}
\begin{document}
\title{Towards an Improved Performance Measure for Language Models}
\author{Joerg Ueberla\\Speech Machines - DERA\\ueberla@signal.dra.hmg.gb\\DERA/CIS/CIS5/TR97426} 

\maketitle

\abstract{
In this paper a first attempt at deriving an improved performance measure for
language models, the probability ratio measure (PRM) is described. 
In a proof of concept experiment, it is shown that PRM correlates better with recognition accuracy and can lead to better recognition results when used as the optimisation criterion of a clustering algorithm.
Inspite of the approximations and limitations of this preliminary work, the results are very encouraging and should justify more work along the same lines.
}

\section{Introduction}
\label{introduction}

The perplexity measure is currently used in the speech recognition and
language modelling community for the following purposes
\begin{itemize}
\item[1)] To evaluate the quality of a language model.	
	\begin{itemize}
	\item[a)] When comparing two language models, the one that has lower perplexity is chosen as the better one.
	\item[b)] When optimising some parameter of a language model (e.g. interpolation weight, discounting parameter, etc.), a parameter value that minimises perplexity is chosen.
	\end{itemize}
\item[2)] To decide on the difficulty of a given recognition task.
\end{itemize}

The main problems with using the perplexity factor for each of these scenarios are
\begin{itemize}
\item[1)] Perplexity is not always well correlated with recognition accuracy. In other words it can happen that language model LM1 has lower perplexity than model LM2, but also has lower recognition accuracy. In the case of 1a) above, this means that a suboptimal language model gets used, thus leading to lower recognition performance. In the case of 1b) above, this means that the language model parameters can get tuned according to a suboptimal criterion.
\item[2)] Two tasks, which have similar perplexities, can nevertheless have very different recognition accuracies.
\end{itemize}

The first of these problems especially is of considerable importance when working on language modeling. There are many modifications to language models that lead to a considerable improvement in perplexity, but to a very small or insignificant increase in recognition accuracy. This implies that full recognition experiments need to be performed before the value of a proposed modification can be properly determined. This can be very time consuming, thereby slowing down the progress in the language modeling field.

The above mentioned problems with the perplexity measure have been known for some time. Yet perplexity is still widely used to evaluate the performance of language models. The main reason for that is that no better, widely accepted alternative measures exist. In this paper, a likely cause of the problems of the perplexity measure will be explored  and directions for alternative measures will be suggested.

\section{Analysis}
\label{analysis}

One possible explanation for the above problems can be described intuitively as follows. The crucial property of a good language model is not that the probability of the correct word is very high in absolute terms, but that it is higher than acoustically confusable words.

This can be further illustrated with the following example. Suppose the task at hand has a vocabulary of four words \{are,bar,cookie,dinner\}. LM1 gives those words the probabilities 0.4, 0.3, 0.2 and 0.1, whereas LM2 assigns 0.1, 0.2, 0.3 and 0.4. Now suppose that, based on acoustic information alone, instances of 'bar' match the acoustic models for 'bar' and 'are' quite well. Then LM1 would be more likely to misrecognise 'bar' as 'are' even though the absolute probability value it assigns to 'bar' (0.3) is higher than that of LM2 (0.2).

This problem is related to a fundamental equation of speech recognition used
to find the best hypothesis. Given the acoustic data $A$, a recogniser chooses
the words sequence $W^{c}$ that is most likely to correspond to $A$, e.g. it
chooses
\begin{equation}
W^{c}=argmax_{W} p(W)*p(A|W). \label{eq:basic}
\end{equation}
Given the distributions $p(W)$ and $p(A|W)$, this is the best choice and leads to 
minimal error rate. However, this equation does not specify how the 
distributions should be estimated. Currently $p(W)$ and $p(A|W)$ are estimated
independently of each other. Ideally, however, one would estimate $p(W)$ and 
$p(A|W)$ together in order to minimise error rate. In theory, one could take account of $p(A|W)$ when estimating $p(W)$ or vice versa. In this paper, however, only the former will be explored.

Let $W^{c}$ denote the correct word sequence. The recogniser will chose the
correct word sequence if 
\begin{equation}
p(W^{c})*p(A|W^{c}) > p(W)*p(A|W) \forall W.  
\end{equation}
Currently, when the language model probabilities $p(W)$ are estimated, one
tries to maximise the likelihood of the training text $p(W_{train})$. One
can argue that if $p(W)$ is large for correct word sequences (like the ones
in the training text), the above equation is more likely to hold and one
is therefore less likely to make errors. 

Following the intuitive explanation of the problems with the perplexity measure, however, one should try to make sure that the likelihood of the correct word sequence is higher than its acoustically confusable alternatives.  
Ideally, one would therefore like to optimise  something like

\begin{equation}
\frac{p(W^{c})*p(A|W^{c})}{\sum_{similar W} p(W)*p(A|W)} \label{eq:simil}
\end{equation}

Equation \ref{eq:simil} can also be derived by saying that one would ideally
want to maximise the a posterior probability

\begin{eqnarray}
p(W^{c}|A)=\frac{p(W^{c})*p(A|W^{c})}{p(A)} \label{eq:posterior}\\
=\frac{p(W^{c})*p(A|W^{c})}{\sum_{W} p(W)*p(A|W)} \nonumber
\end{eqnarray}

If $p(A|W)$ is assumed to be negligible for acoustically dissimilar $W$, this
can be approximated by equation \ref{eq:simil} above.

\section{Experiments and Results}
\label{Results}

In order to explore the ideas mentioned above, several experiments were
performed on an Airborne Reconnaissance Mission (ARM) task, which has a vocabulary
of about 500 words. These experiments are aimed at the problems 1a) and 1b)
mentioned above and are only intended as a proof of concept. In order to
minimise the amount of coding necessary, only bigram experiments were
performed. Furthermore, rather than looking at confusable word sequences
in general, only one-to-one substitutions were considered.
Thus, equation \ref{eq:simil} was further approximated as follows.

Let $W^{c}$ and $W$ be two word sequences which only differ in word $j$, e.g. $w^{c}_{i}=w_{i}$ except for $i=j$. The ratio
of their probabilities is

\begin{eqnarray}
\lefteqn{\frac{p(W^{c})*p(A|W^{c})}{p(W)*p(A|W)}=} \\
&=& \frac{ \prod_{i} p(w^{c}_{i}|w^{c}_{i-1}) * p(A|W^{c})}{\prod_{i} p(w_{i}|w_{i-1})*p(A|W)} \nonumber \\
&=& \frac{ p(w^{c}_{j}|w^{c}_{j-1})*p(w^{c}_{j+1}|w^{c}_{j}) * p(A|W^{c})}{ p(w_{j}|w^{c}_{j-1})*p(w^{c}_{j+1}|w_{j})*p(A|W)} \nonumber
\end{eqnarray}

If we further approximate the ratio of the acoustic probabilities of the two word sequences by a similarity measure $Sim(w^{c}_{j},w_{j})$, one obtains

\begin{eqnarray}
\lefteqn{\frac{p(W^{c})*p(A|W^{c})}{p(W)*p(A|W)}=} \\
&=& \frac{ p(w^{c}_{j}|w^{c}_{j-1})*p(w^{c}_{j+1}|w^{c}_{j})}{ p(w_{j}|w^{c}_{j-1})*p(w^{c}_{j+1}|w_{j})} * Sim(w^{c}_{j},w_{j}) \nonumber
\end{eqnarray}

Using this as the basic component, the average, normalised ratio of probabilities of the correct word over its confusable alternatives is 

\begin{equation}
\prod_{similar w_{j}} \frac{p(w^{c}_{j}|w^{c}_{j-1})*p(w^{c}_{j+1}|w^{c}_{j})}{p(w_{j}|w^{c}_{j-1})*p(w^{c}_{j+1}|w_{j})}) *Sim(w^{c}_{j},w_{j}))^{1/|similar w_{j}|}
\end{equation}

Extending this to substitutions at any point in the correct word sequence, the probability ratio measure, PRM, is calculated as 

\begin{equation}
\prod_{w^{c}_{i}} (\prod_{similar w_{i}} \frac{p(w^{c}_{i}|w^{c}_{i-1})*p(w^{c}_{i+1}|w^{c}_{i})}{p(w_{i}|w^{c}_{i-1})*p(w^{c}_{i+1}|w_{i})}) *Sim(w^{c}_{i},w_{i}))^{1/|similar w_{i}|}
\end{equation}

By rearranging terms, this can be further rewritten as

\begin{equation}
\prod_{w^{c}_{i}} (\prod_{similar w_{i}} \frac{p(w^{c}_{i}|w^{c}_{i-1})*Sim(w^{c}_{i},w_{i})}{p(w_{i}|w^{c}_{i-1})})^{1/|similar w_{i}|} (\prod_{similar w_{i-1}} \frac{p(w^{c}_{i}|w^{c}_{i-1})}{p(w^{c}_{i}|w_{i-1})}))^{1/|similar w_{i-1}|} \label{eq:used}
\end{equation}

The similarity measure $Sim(w^{c}_{i}, w_{i})$ was calculated based on the
human equivalent noise ratio (HENR) measure as described in \cite{Moo77}. The number of acoustically confusable words $NbSimil$ was varied from $0$ (in which case the normal perplexity measure is calculated) to $80$.

In a first experiment, recognition experiments were run on 30 testing utterances.
The standard perplexity measure, and the probability ratio measure (PRM) given by equation \ref{eq:used} were also calculated on each of these files. The correlation between the 
recognition accuracy and the different measures were then calculated using Spearmans
rank order correlation coefficient $r_{S}$ (from the numerical recipes book \cite{Pre88}, pp.507ff).  It essentially calculates the linear correlation coefficient of the set of points where the actual sample values are replaced by their rank among all the samples. 
The results
are shown in Table \ref{tab:correlation}, for different numbers of similar words considered.

\begin{table}
\centering
\begin{tabular}{|c|c|} \hline
NbSimil & Correlation Coefficient $r_{S}$ \\  \hline
0 (e.g. perplexity)	& -0.42 \\
10			& 0.47 \\
20			& 0.50 \\
40			& 0.46 \\
80			& 0.45 \\ \hline
\end{tabular}
\caption{Correlation coefficient $r_{S}$ of recognition accuracies and the probability ratio measure on thirty utterances for various numbers of acoustically confusable words considered}
\label{tab:correlation}
\end{table} 

Lower perplexities do in general lead to higher accuracy, which explains the 
negative value of the correlation coefficient for $NbSimil=0$, e.g. the normal perplexity measure. On the other hand, higher 
ratio measure values lead in general to higher accuracies, which is why
the corresponding correlation coefficients are positive for all other entries, corresponding to various parameter settings of the PRM measure. When looking at
the absolute value of the correlation coefficients, which is the actual value of importance, one can see that the
ratio measure is more correlated than the perplexity measure.

%\fix{calculate PP and PRM measure for the 4 different LMs in talbe 2 for which I have recognition figures; this would be the proper thing, rather than correlation at a file level}

%\fix{use new measure for different HENR clustering iteration (PP and Acc seems to vary widely)}

%\fix{calculate PRM on 93h1p0 files, for which I already have Acc and PP per file}

In a second experiment, equation \ref{eq:used} was used as optimisation
function of a clustering algorithm (instead of the usual one, which is
derived from the perplexity). This is done in a manner similar to 
\cite{Ueb97a}, where a modified optimisation criterion is derived for
a different purpose.
The results
are shown in Table \ref{tab:accuracy}, for different numbers of similar words considered.
Depending on the number of similar words used, recognition results that are
better than those using the normal optimisation criterion can be obtained.

\begin{table}[b]
\centering
\begin{tabular}{|c|c|} \hline
NbSimil & Recognition Accuracy (in \%)\\  \hline
0 (e.g. perplexity)	& 87.1 \\
10			& 85.9 \\
20			& 87.5 \\
40			& 87.2 \\ \hline
\end{tabular}
\caption{Recognition accuracy for clustered language models built with optimisation functions derived from the probability ratio measure for various numbers of acoustically confusable words considered }
\label{tab:accuracy}
\end{table} 

\section{Conclusions and Future Work}
\label{conclusions}

In this paper a first attempt at deriving an improved performance measure for
language models, the probability ratio measure (PRM) is described. 
In a proof of concept experiment, it is shown that PRM correlates better with recognition accuracy and can lead to better recognition results when used as the optimisation criterion of a clustering algorithm.
However, it is intended as a proof of concept only and
therefore has many limitations. The acoustic similarity measure used, for
example, is not derived from actual Hidden Markov Models. Moreover, only pairwise
confusions rather than complete confusable paths with substitutions and
insertions were considered. In spite of these severe approximations, the results are very encouraging and should justify more
work along the same lines, possibly addressing the limitations mentioned
above. 

%\bibliography{/home/ueberla/latex/inputs/all,/home/ueberla/latex/inputs/alphabetical}
\bibliographystyle{plain}

\end{document}